\documentclass[letterpaper, 10 pt, conference]{ieeeconf}  

\IEEEoverridecommandlockouts                              

\usepackage{cite}
\usepackage{amsmath}
\usepackage{siunitx}
\usepackage{graphicx}
\usepackage{booktabs}
\usepackage{multirow}

\DeclareMathOperator*{\argmax}{arg\,max}

\usepackage{moreverb,url}
\usepackage[colorlinks,bookmarksopen,bookmarksnumbered,citecolor=black,urlcolor=blue,linkcolor=black]{hyperref}

\title{\LARGE \bf
MPGNet: Learning Move-Push-Grasping Synergy 

for Target-Oriented Grasping in Occluded Scenes}
\author{Dayou Li, Chenkun Zhao, Shuo Yang, Ran Song, Xiaolei Li, and Wei Zhang${^*}$
\thanks{All the authors are with the School of Control Science and Engineering, Shandong University, Jinan, 250061, China.}
\thanks{*Corresponding author: Wei Zhang (email: davidzhang@sdu.edu.cn)}
}

\begin{document}

\maketitle

\thispagestyle{empty}
\pagestyle{empty}

\begin{abstract}
This paper focuses on target-oriented grasping in occluded scenes, where the target object is specified by a binary mask and the goal is to grasp the target object with as few robotic manipulations as possible. Most existing methods rely on a push-grasping synergy to complete this task. To deliver a more powerful target-oriented grasping pipeline, we present MPGNet, a three-branch network for learning a synergy between moving, pushing, and grasping actions. We also propose a multi-stage training strategy to train the MPGNet which contains three policy networks corresponding to the three actions. The effectiveness of our method is demonstrated via both simulated and real-world experiments. Video of the real-world experiments is at \url{https://youtu.be/S_QKZqkh0w8}.

\end{abstract}

\section{Introduction}


Grasping is a foundational action for versatile robotic tasks. Researchers have proposed many methods \cite{sahbani2012overview,mahler2017dex,asif2018ensemblenet,lenz2015deep,fang2020graspnet,li2023mobile,berscheid2019robot,zeng2018learning,huang2021dipn,deng2019deep} for robotic grasping, which usually leverage deep learning and/or reinforcement learning to generate grasping policies and show good performance in scenes without occlusion. However, in real-world grasping scenarios, objects are often randomly placed in occluded scenes, bringing difficulties to grasping action. To solve this problem, many researchers have studied pre-grasp synergy, most of which extend non-prehensile pushing action to action space. Some methods  \cite{berscheid2019robot,zeng2018learning,huang2021dipn,deng2019deep} combined pushing and grasping policies for sequential manipulation as pushing can help rearrange the objects, making it easier for performing grasping. Such methods focus on target-agnostic grasping tasks, and cannot perform grasping according to user specification.



Target-oriented grasping aims to grasp user-specified objects. It is particularly challenging for occluded scenes. This is because it is non-trivial to accurately find and grasp the target object based only on partial observations. Most existing methods \cite{xu2021efficient,zuo2023graph,yang2020deep,yu2023iosg,li2022learning} for target-oriented grasping used a two-branch Q-network architecture to learn the push-grasping synergy to facilitate target-oriented grasping in occluded scenes. For example, Xu et al. \cite{xu2021efficient} combine RGB-D heightmap and goal segmentation mask as input and train two separate Q-networks to learn the synergy between pushing and grasping actions. Instead of using goal segmentation masks, Yu et al. \cite{yu2023iosg} design a target similarity network (TSN) to estimate goal localization. They also pass the visual feature into two deep Q-networks for making pushing-or-grasping decisions. Li et al. \cite{li2022learning} adopt a similar two-branch deep Q-network to estimate the Q value distribution while decoupling the position and angle predictions.

\begin{figure}[t]
\setlength{\abovecaptionskip}{-0.1cm}
\begin{center}
\includegraphics[width=0.9\linewidth]{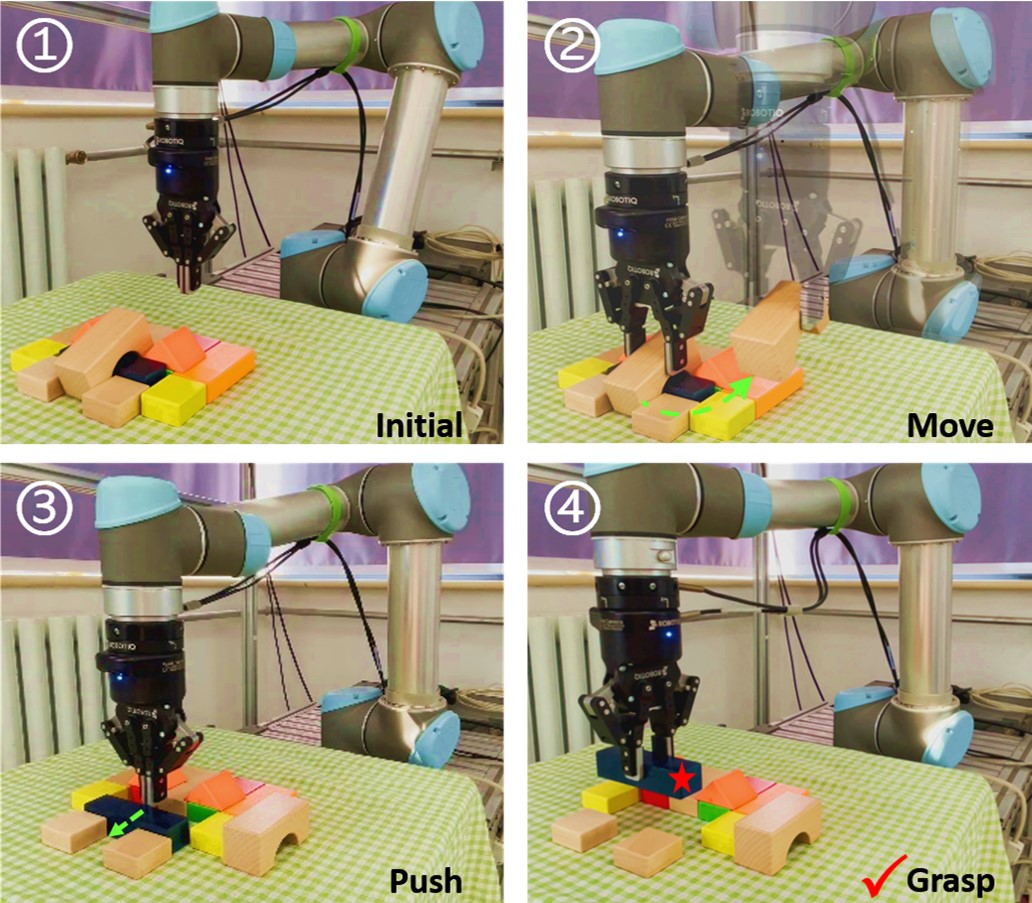}
\end{center}
\caption{Through the synergy of moving, pushing, and grasping actions, the robotic arm based on the proposed MPGNet can efficiently grasp the target object in occluded scenes.}
\label{intro}
\end{figure}
One important assumption made by the above methods is that the agent should select the action (pushing or grasping) with a larger Q value at each training step. However, this assumption has poor interpretability and is not always reliable. Besides, most existing methods addressed the challenges of grasping in occluded scenes by pushing the stacked objects in specific directions while the pushing actions sometimes cannot efficiently solve the target-oriented grasping task, and even further mess up the grasping scenarios. 

In this paper, we rethink the pipeline for addressing the target-oriented grasping in occluded scenes. First, we design three primitive actions, moving, pushing, and grasping, which work together to complete the tasks. Second, instead of solely selecting actions based on the scale of the Q value, we present a human intuition-guided workflow to coordinate the three primitive actions. Specifically, we design a three-branch deep Q-network to learn a synergic policy for target-oriented grasping in an occluded scene. Since making use of three primitive actions inevitably increases the learning difficulty compared to previous methods based on the push-grasping synergy, we design a multi-stage training strategy to deliver stable and fast policy training. Fig.~\ref{intro} illustrates the core idea of this work through a real-world example. 


To summarize, the main contributions of this paper are:
\begin{itemize}
\item We propose a novel three-branch deep Q-network, namely MPGNet, which learns a synergy between moving, pushing, and grasping actions for target-oriented grasping in occluded scenes. 
\item We design a multi-stage strategy to train the MPGNet where we first train each action separately and then carry out a joint training to learn the move-push-grasping synergy effectively.
\item We perform extensive experiments in the simulation to evaluate MPGNet, and further demonstrate a real-world robotic system that can reliably complete the target-oriented grasping tasks in occluded scenes.
\end{itemize}

\section{Related Work}
\subsection{Target-Oriented Grasping without Action Synergy}
Some scholars propose to obtain the target object via a single grasping action\cite{murali20206,li2023mobile,liri2021real,zeng2022robotic,lin2022know,chen2021joint,yang2021learning}. We classify those methods as two-stage and one-stage. The two-stage implies that the target specification and grasp detection are not fused, but rather that task-agnostic grasp detection is first performed, followed by target filtering using a suitable heuristic \cite{murali20206,li2023mobile,liri2021real,zeng2022robotic}. Murali et al. \cite{murali20206} crop the point cloud of the scene using a binary mask of the target object and then perform 6-Dof grasp detection on the target object point cloud. Liri et al. \cite{liri2021real} use voice to specify the target object from object detection results and then forward it to the grasp detection module.
An alternative, denoted as one-stage, directly generates grasp proposals based on the input target information in the form of images and language instructions \cite{lin2022know,chen2021joint,yang2021learning}. Lin et al. \cite{lin2022know} propose to use sketches to represent the target objects and designed an end-to-end network for learning to generate target-oriented grasps. Yang et al. \cite{yang2021learning} design a multi-object dense descriptor for learning target-oriented grasp affordance via DRL with the targets defined by RGB images.

However, the above methods perform poorly in occluded scenes due to the limited action space.  Obtaining the target object with a single grasping action is virtually impossible in some occluded and cluttered scenes. 
Therefore, it is necessary to explore action synergy algorithms to address the target-oriented grasping tasks in occluded scenes.


\subsection{Target-Oriented Grasping with Action Synergy}
Many studies explore the synergy of pre-grasp manipulations such as pushing and grasping to improve the success rate of grasping the target \cite{xu2021efficient,zuo2023graph,yang2020deep,yu2023iosg,li2022learning}.
Xu et al. \cite{xu2021efficient} propose a goal-conditioned hierarchical reinforcement learning formulation including goal relabeling strategy and alternative training to handle sample inefficiency and accelerate training. Yang et al. \cite{yang2020deep} propose a Bayesian-based policy to explore the target along with a classifier-based policy for push-grasping synergy. In order to solve the learning problems in large state space, Li et al. \cite{li2022learning} decouple the action space by learning the position and the angle of the grasp in a separate way, which means that an additional branch is used to predict grasp angles. Moreover, some new methods are also gradually coming out. In \cite{zuo2023graph}, Zuo et al. adopt a graph-based deep reinforcement learning model instead of classical DQN to explore invisible objects so as to achieve better performance for push-grasping synergy.

Currently, the existing action synergy methods for target-oriented robotic grasping are mainly based on pushing and grasping. However, from our experimental test of the methods \cite{xu2021efficient,yang2020deep,zeng2018learning}, we discover that in some common occlusion situations (as shown in Fig.~\ref{intro}), push-grasping synergy is not able to handle it efficiently. Firstly, for stacked occlusion, it is difficult for the agent to perform effective pushing actions, which is determined by the definition of the pushing action. Secondly, for push-grasping synergy, in some occasions, multiple action steps are required to successfully grasp the target object, which is time-consuming. Inspired by how humans deal with such occlusion, we expand the action space dimension by building a three-branch action synergy network. It is worth mentioning that Liu et al. propose GE-Grasp \cite{9981499} to utilize nontarget-oriented grasps to reduce occlusion, however, it involves large amounts of data collection and fails to provide an effective framework to learn the synergy between different actions. Instead, we propose an automatic pipeline to handle this problem by adopting hierarchical reinforcement learning. 


\begin{figure*}[t]
\begin{center}
\includegraphics[width=1\linewidth]{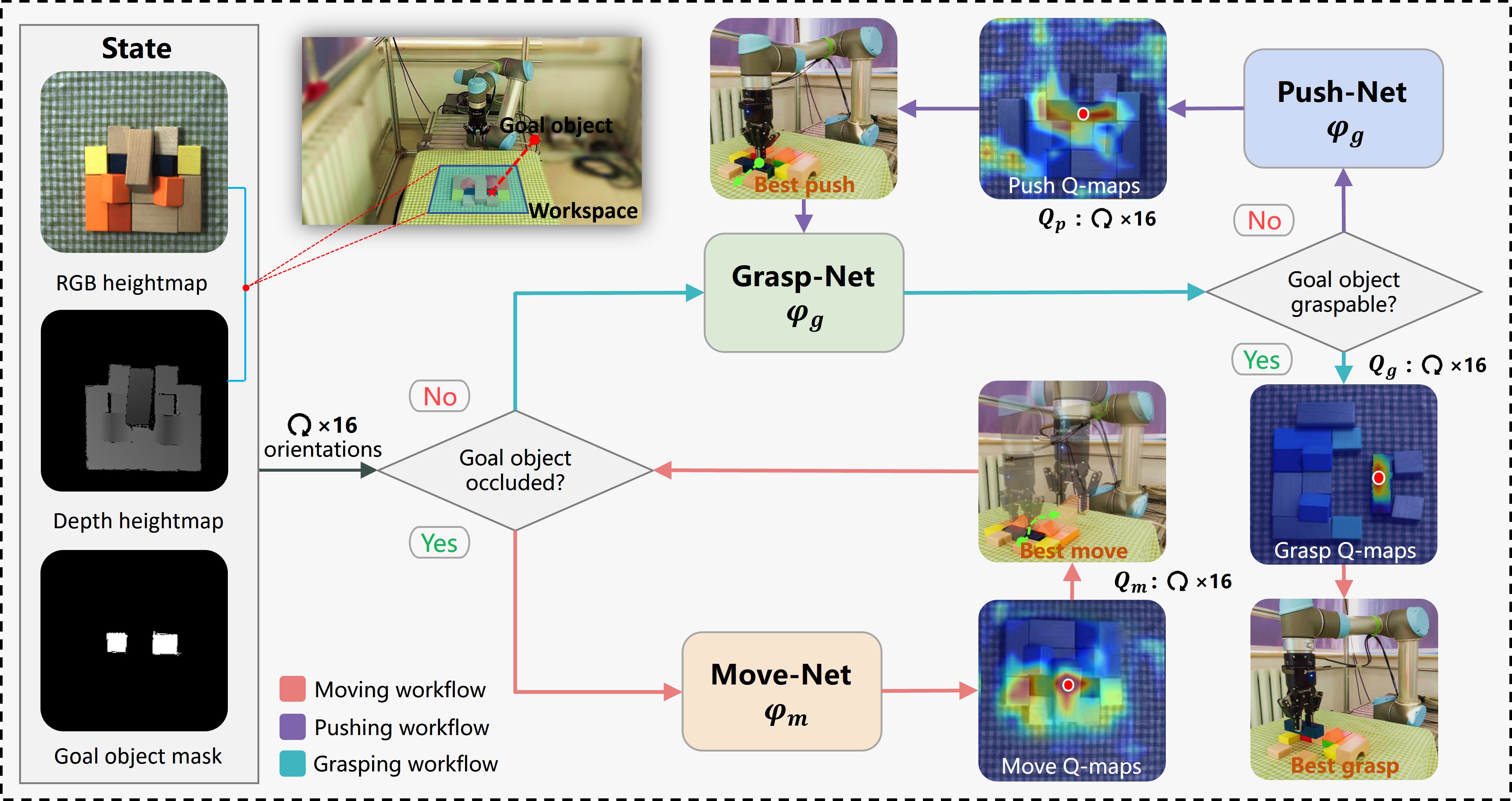}
\end{center}
\caption{Overview of MPGNet. The input data of MPGNet are obtained by an RGB-D camera from a top-down view. The heightmaps are rotated at 16 angles to predict different motion orientations and fed into MPGNet. Move-net, grasp-net, and push-net work collaboratively to grasp target objects in occluded scenes. The move-net is designed to remove occluding objects to make the target as graspable as possible. When there is no occlusion in the workspace, the pushing action assists in grasping the target object.}
\label{network}
\end{figure*}
\section{Method}
Fig. \ref{network} illustrates the pipeline of the proposed MPGNet. In the following, we elaborate each of its components. 
\subsection{Problem Statement} 

We define the target-oriented grasping task within the framework of a goal-conditioned Markov decision process. At any time step \( t \), given the current state \( s_t \) and goal mask \( g_t \) for a specified target object, the agent must select and perform an action \( a_t \) based on the policy \( \pi(s_t | g_t) \).
Subsequently, the system transitions to a new state $s_{t+1}$ and receives an immediate reward $R(s_{t},a_{t},s_{t+1}|g_{t})$. The objective of our deep reinforcement learning task is to discover the optimal policy $\pi^{*}$ that maximizes the expected cumulative future rewards, represented as:
\begin{equation}
    R_{t}=\sum_{i=t}^{\infty}{\gamma^{i-t}R(s_{i},a_{i},s_{i+1}|g_{i})}
\end{equation}
The discount factor is denoted by $\gamma$. In this study, we explore the use of deep Q-learning to train neural networks that approximate the action-value function $Q_{\pi}(s_{t},a_{t}|g_{t})$.
The expected reward for taking action $a_{t}$ in the state $s_{t}$ at time $t$ is evaluated. A greedy deterministic policy $\pi(s_{t}|g_{t})$ is derived by choosing the action based on the maximum Q values. Moreover, our learning objective is to minimize the temporal difference error $\delta_{t}$ of $Q(s_{t},a_{t}|g_{t})$ compared to a target value $y_{t}$:
\begin{equation}
    \delta_{t} = |Q(s_{t},a_{t}|g_{t})-y_{t}|
\end{equation}
\begin{equation}
    y_{t} = R(s_{t},a_{t},s_{t+1}|g_{t}) + \gamma Q(s_{t+1},\argmax_{a}(Q(s_{t+1},a|g_{t})))
\end{equation}
where $a$ is the total set of available actions.
This paper focuses on occluded scenes where the target object is occluded by other objects from a heightmap perspective.

\subsection{Three-Branch Deep Q-Network}
\subsubsection{State Representations}
 To better represent each state $s_{t}$, the RGB images and the depth images captured by a fixed camera are orthographically back-projected upwards in the gravity direction to generate the color heightmap $c_{t}$ and the depth heightmap $d_{t}$. The state of the workspace at time $t$ is represented as $s_{t}=(c_{t},d_{t})$. We use a mask heightmap to represent the target object as $g_{t}$, which indicates the area occupied by the target object in the workspace. 
In the simulated environment, the mask heightmap is obtained through the following steps: 1) We obtain the position and shape information of the target object using APIs provided by the simulator and then calculate the area occupied by the target object along the direction of gravity. 
2) We populate the mask heightmap with values derived from the spatial data of the target object's occupied area, ensuring the heightmap reflects only the general boundaries of the target object.
 

\subsubsection{Action Space}
We define each action \( a_t \in \{a_m, a_g, a_p\} \) as a distinct motion primitive, where \( a_m \), \( a_g \), and \( a_p \) correspond to the behaviors of moving, grasping, and pushing, respectively. The specifics of each motion primitive are detailed below.

\textbf{Moving.}
A moving action is designed to move the obstacles, defined as $a_{m}=(p_{m},\theta_{m})$.
Let $p_{m}=(x_{m},y_{m},z_{m}) \in \mathbf{R}^{3}$ denote the central position of the two-finger robotic gripper within the workspace, where $\theta_{m} \in \mathbf{R}$ signifies the rotational angle. The coordinates $(x_{m},y_{m})$ are derived from the pixel location in $Q_{m}$ through eye-to-hand calibration. The $z_{m}$ coordinate is defined as $z_{m}=h_{m}-4cm$, with $h_{m}$ representing the height at the point $(x_{m},y_{m})$. During operation, the gripper must descend 4 cm below $h_{m}$ before the fingers close. The angle $\theta_{m}$ is one of 16 possible orientations around the z-axis, ranging from $0\si{\degree}$ to $360\si{\degree}$, with a $22.5\si{\degree}$ increment between each orientation.


\textbf{Pushing.}
A pushing motion executed by the gripper's tip is represented as $a_{p}=(p_{p},d_{p})$. Each push has a consistent length of 10 cm, following a straight-line trajectory. The starting point of the push is denoted by $p_{p}=(x_{p},y_{p},z_{p}) \in \mathbf{R}^{3}$, where $d_{p} \in \mathbf{R}^{3}$ specifies the direction of the push. The coordinates $(x_{p},y_{p})$ are derived from pixel data in $Q_{p}$ through calibration. To ensure the gripper lightly contacts the object, we set $z_{p}=h_{p}-1 cm$ if $h_{p}$ exceeds 0. Conversely, when $h_{p}$ is negative, $z_{p}$ is adjusted to 2 cm to prevent the gripper from touching the ground. The push direction is one of 16 possible orientations within a $0\si{\degree}$ to $360\si{\degree}$ range around the z-axis.

\textbf{Grasping.}
A grasping action can be described as \( a_{g} = (p_{g}, \theta_{g}) \), where \( p_{g} = (x_{g}, y_{g}, z_{g}) \in \mathbf{R}^{3} \) indicates the center point of the top-down parallel-jaw grasp, and \( \theta_{g} \in \mathbf{R} \) denotes the grasping angle, which varies from \( 0\si{\degree} \) to \( 360\si{\degree} \) around the z-axis, divided into 16 segments. The height at the position \( (x_{g}, y_{g}) \) is represented by \( h_{g} \), and the value of \( z_{g} \) is calculated as \( h_{g} - 4 \text{cm} \). To perform the grasp, the gripper must descend 4 cm below \( h_{g} \).

\subsubsection{Network Details}We extend vanilla deep Q-networks \cite{mnih2015human} by designing our own Q-functions as three feed-forward fully convolutional networks (FCNs). We name it MPGNet which includes move-net $\varphi_{m}$, push-net $\varphi_{p}$, and grasp-net $\varphi_{g}$. MPGNet takes as input the state representations $s_{t}=(c_{t},d_{t})$ with goal $g_{t}$ and outputs dense pixel-wise Q-value maps $Q=(Q_{m},Q_{g},Q_{p}) \in \mathbf{R}^{224\times224\times3}$ with the same resolution of $s_{t}$, where $Q_{m},Q_{g},Q_{p} \in \mathbf{R}^{224\times224}$ represent Q-values of the corresponding actions in the pixel positions. For each net, we adopt two parallel MobileNetV3 \cite{howard2019searching} pre-trained on ImageNet to encode color heightmaps and depth heightmaps. A public MobileNetV3 block is used for feature extraction of the goal mask heightmaps. 
Prior to being input into MobileNetV3, each heightmap undergoes rotation across 16 different angles, facilitating the learning process for motion primitives related to movement, grasping, and pushing actions. The features extracted are initially concatenated along the channel dimension and then processed through two $1\times1$ convolutional layers. These layers incorporate nonlinear activation functions (ReLU) and are followed by spatial batch normalization. The final step involves bilinear upsampling. The entire network produces 48 pixel-wise Q-value maps, divided into 16 maps for movement at various orientations, 16 maps for pushing in different directions, and 16 maps for grasping at various angles
The goal of the moving action is to make enough space around the target object for successive grasping by clearing the occluding objects that the pushing action cannot handle efficiently. 
We restrict the movement to the vicinity of the target object, allowing the network to concentrate solely on this area. This approach ensures that the movement directly addresses any occlusion affecting the target object.

\subsection{Policy Learning}
We model the move-push-grasping synergy for target-oriented grasping as a hierarchical reinforcement learning problem. We set the grasp-net $\varphi_{g}$ as the discriminator to score each state $s_{t}$ for grasping the target object. In the early stage of training the move-net $\varphi_{m}$, it cannot learn how to handle occlusion properly, which causes negative rewards and leads to inefficient learning. Thus, when occlusion occurs, the well-trained grasp-net $\varphi_{g}$ is used to score the state $s_{t}$ changed by the moving action. If the increase of improved grasp Q-values exceeds a threshold, the agent will still receive a positive reward regardless of whether the occlusion is removed by the moving action. 
When occlusion is absent, pushing actions are initiated, and the robot performs these actions to alter the object's state until it becomes suitable for grasping by the discriminator.
We design a multi-stage strategy to train MPGNet. The details are as follows.

\textbf{Stage I: Target-Agnostic Grasping-Only Training.} 
Since we use the grasp-net $\varphi_{g}$ to score the states, we need to ensure that it is well-trained to accurately find the stable grasping positions. 
At this phase, the robot executes the grasping task without targeting specific objects, which means that the input mask should include all the objects in the scene. 
The episode concludes once every item in the workspace has been successfully grasped.
To reduce the effect of occlusion, we make the locations where $m$ objects are dropped relatively discretely. We define the target-agnostic grasping reward function as:
\begin{equation}
    R_{g1} = \left \{\begin{aligned}
        & 1,\quad \textrm{grasping a object successfully} \\
        & 0,\quad \textrm{otherwise}
    \end{aligned}
            \right.
\end{equation}

\textbf{Stage II: Target-Oriented Grasping-Only Training.} 
In the previous stage, we obtain a relatively stable grasp net for target-agnostic use. In this stage, we aim to train a target-oriented grasp net based on the previous stage. In practice, we randomly drop $m$ objects in the workspace and set the $n$-th object as our target. For learning efficiently, we make the target object relatively isolated from other objects. We define the target-oriented grasping-only reward function as:
\begin{equation}
    R_{g2} = \left \{\begin{aligned}
        & 1,\quad \textrm{grasping the target successfully} \\
        & 0,\quad \textrm{otherwise}
    \end{aligned}
            \right.
\end{equation}

Obviously, there are two occasions where the agent receives zero reward: failed grasps and wrong grasps. 
After undergoing a substantial number of training episodes, the $Q$ values associated with successful grasp attempts tend to converge to a particular value, denoted as $Q_{g}^{*}$. This threshold indicates that when a state's $Q$ value surpasses this point, the target object is likely graspable. Subsequently, in later stages, the grasp network $\varphi_{g}$ from stage II functions as a discriminator.

\textbf{Stage III: Target-Oriented Moving-Only Training.} In this stage, the moving policy is learned to clear occlusions for the target object. To obtain the status of whether the target object is occluded, we propose an occlusion-checking mechanism in the simulation environment: 1) Crop the depth heightmap of the target object through the given target mask. 2) Obtain the position and shape information of the target object via the simulation API. 3) Calculate the highest (z-axis) point on the surface of the target object, and compare the calculated highest point with the highest z value on the cropped depth heightmap to generate the occlusion status. To train the move net, we create some simple occluded scenes at first: drop two objects in sequence and ensure by setting two positions adjacent that the first object is occluded by the other one. We set a flag of moving successfully: after a moving action, if the occlusion of the target object is removed and the target is not grasped by the moving action, the flag is set to true. Consequently, the agent should receive a positive reward. Each episode ends when the flag of moving successfully turns from false to true. We set the maximum threshold as 5 for the number of moving actions, which means that the simulation environment will be reset if the agent executes 5 moving actions without clearing occluding objects. After several training episodes, the agent can easily handle occluded scenes. Next, we add more objects in the workspace to make it more scattered for the agent to learn to handle more occluded scenarios.
In the early training stage of occluded scenes, it is usually difficult for the agent to meet the conditions of moving successfully. 
Thus we give rewards by introducing the method of comparing $Q_{g}$ before and after moving. The target-oriented moving-only reward function is defined as:
\begin{equation}
    R_{m} = \left \{\begin{aligned}
        & 1,\quad \textrm{move successfully} \\
        & 0.5,\quad Q^{\textrm{im}}_{g}>0.5 \quad\textrm{and change detected} \\
        & -0.5, \textrm{ move the target or no change detected}
    \end{aligned}
            \right.
\end{equation}
where `$\textrm{move the target}$' denotes that the target object is grasped by the moving action, and `$\textrm{change detected}$' denotes that the surrounding of the target object is changed. $Q^{\textrm{im}}_{g}=Q_{g}^{\textrm{am}}-Q_{g}^{\textrm{bm}}$, where $Q^{\textrm{im}}_{g}$ means the improved grasp Q-value by a moving action, $Q_{g}^{\textrm{am}}$ means the grasp Q-value after a moving action, $Q_{g}^{\textrm{bm}}$ means the grasp Q-value before a moving action.

\textbf{Stage IV: Target-Oriented Pushing-Only Training.}
In this stage, our primary objective is to train a push network, while keeping the weights of the pre-trained grasp network and move network unchanged. Additionally, the grasp network functions as a discriminator to evaluate and score the pushing actions.
Positive rewards will be given to the agent if the pushing actions improve the Q values predicted by the grasp net.
In practice, the number of pushing actions is limited to 5 in an episode with a grasp at the end. Each pushing action aims to continuously increase $Q_{g}$ of the target object until it exceeds $Q_{g}^{*}$. We construct some scenes where the target objects are surrounded by other objects.
The target-oriented pushing-only reward function is defined as:
\begin{equation}
    R_{p} = \left \{\begin{aligned}
        & 0.5,\quad \textrm{if} \quad Q^{\textrm{ip}}_{g}>0.1 \quad \textrm{and change detected} \\
        & -0.5, \quad \textrm{no change detected} \\
        & 0,\quad \textrm{otherwise}
    \end{aligned}
            \right.
\end{equation}   
where $Q^{\textrm{ip}}_{g}=Q_{g}^{\textrm{ap}}-Q_{g}^{\textrm{bp}}$. 
$Q^{\textrm{ip}}_{g}$ means the improved grasp Q-value by a pushing action, $Q_{g}^{\textrm{ap}}$ means the grasp Q-value after a pushing action, $Q_{g}^{\textrm{bp}}$ means the grasp Q-value before a pushing action.

\textbf{Stage V: Target-Oriented Joint Training.} 
In earlier stages, our grasp net was trained exclusively on grasping actions within a cluttered environment, while our move net was trained in a specifically designed scenario with occlusions. Additionally, our push net was trained in a moderately occluded setting where push-grasp synergy was emphasized.
However, when combining them, there will be a problem of distribution mismatch as the training prerequisites differ for each network. Moreover, the move-push-grasping synergy needs further training to better suit real scenarios. 

To solve this problem, we further train the move-push-grasping policy based on the previously trained networks $\varphi_{m}$, $\varphi_{g}$, and $\varphi_{p}$. In this stage, all three policies are alternatively trained with other nets' weights frozen. Through this stage, the three actions coordinate with each other better.
We set the episode the same for moving, grasping, and pushing as in previous stages. 10 objects are randomly dropped in the workspace to construct a realistic and occluded scene. In each episode, the agent first checks the occlusion status to decide whether moving actions should be executed. Then, the pushing actions are executed till the maximal Q value of the target object exceeds the threshold $Q^{*}_{g}$. The reward functions for each action remain the same as in the previous stages. 

\begin{figure}[t]
\begin{center}
\includegraphics[width=1\linewidth]{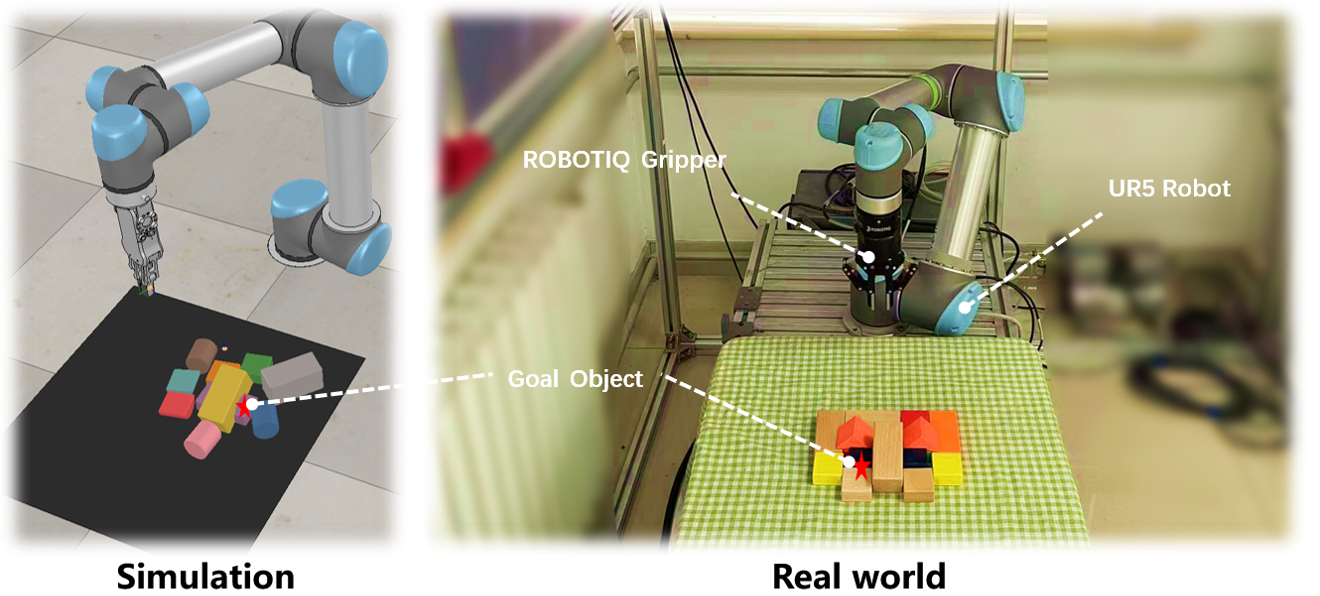}
\end{center}
\caption{We train MPGNet in the simulated environment and then transfer it to the real world.}
\label{system}
\end{figure}

\subsection{Training Details}
In the first two stages of the grasping-only training, we set the number of objects in the scene as $m=5$. To determine $Q_{g}^{*}$ for the discriminator, we adopt the methods in \cite{xu2021efficient}. After we finish the first two stages, the Q value becomes stable around $Q_{g}^{*}=1.85$. Thus we set $Q_{g}^{*}=1.85$ as the threshold to determine whether the target object is graspable. Considering that the grasping and the moving positions should be close to the target, we propose to generate the rough target mask by calculating the largest bounding rectangle of the target object, and assign the Q value out of the rough mask to 0. The loss function in our method is the Huber loss function defined as:
\begin{equation}
    L = \left \{\begin{aligned}
        & \frac{1}{2}\delta_{t}^{2},\quad if\;\delta_{t}<1 \\
        & |\delta_{t}-\frac{1}{2}|,\quad \textrm{otherwise}
    \end{aligned}
            \right.
\end{equation}
where $\delta_{t}$ is defined in Section III-A. We also adopt $\epsilon$-greedy exploration strategy with $\epsilon$ initialized as 0.5 and annealed to 0.1. We set future discount $\gamma=0.5$ as constant. The network is trained with Adam optimizer with a fixed learning rate $10^{-4}$, a weight decay $2^{-5}$, and betas $(0.9,0.99)$.

\section{Experiments}




As shown in Fig.~\ref{system}, we train MPGNet in the simulated environment and then transfer it to the real world. In our simulated environment (CoppeliaSim), a UR5 robotic arm with an RG2 gripper is mounted in the workspace. Two cameras are mounted at fixed positions with known extrinsic parameters to capture RGB-D images of the workspace. In the real world, we build up a scene similar to the simulation: we use a UR5 robotic arm equipped with a ROBOTIQ two-finger gripper as the agent to execute different actions, and a RealSense D435i RGB-D camera is fixed horizontally above the workspace to capture color and depth images. 
In the following, we evaluate the proposed method via a series of experiments. 

\subsection{Baselines}
We compare the performance of our method with the following baselines:

\textbf{Grasping the Invisible} \cite{yang2020deep} is a target-oriented approach that uses a classifier-based policy to coordinate pushing and grasping actions to grasp the target object in clutter. A color segmentation method is used to generate target masks and to detect whether the target is visible for coordination or exploration. 

\textbf{Efficient Push-Grasping} \cite{xu2021efficient} uses several techniques to train a target-oriented push-grasping synergy. Goal relabeling is adopted to improve sample efficiency. 
Moreover, alternative training is applied in the training process which means that pushing and grasping are learned in turn.    

\textbf{Grasping-Only} applies a single FCN network to learn a greedy deterministic grasping policy. 

\textbf{Target-Oriented VPG} is a target-oriented approach that extends VPG \cite{zeng2018learning} by taking the target mask in addition as input to learn the push-grasping synergy. VPG outputs Q maps of both pushing and grasping for target-agnostic tasks. It executes the action with the highest Q value in the target area.


\subsection{Evaluation Metrics}
We evaluate our method in some scenes with occlusion where $n$ objects are randomly dropped at relatively adjacent positions in the simulation environment. We set the first or the second object as the target object to ensure that the target is occluded. The metrics are defined as below:
\subsubsection{Task Success Rate}
In $n$ test experiments, the task success rate is defined as the average percentage of successful tests over $n$ runs. A test is considered successful if the agent grasps the target object successfully within 10 motions.

\begin{figure}[t]
\setlength{\abovecaptionskip}{-0.1cm}
\begin{center}
\includegraphics[width=1\linewidth]{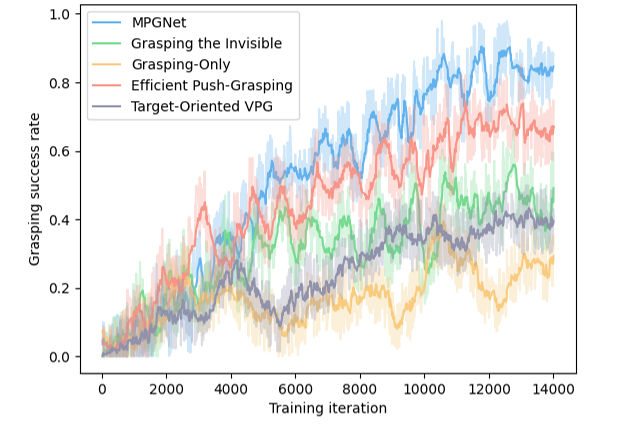}
\end{center}
    \caption{Learning curves of different methods.}
\label{training performance}
\end{figure}

\begin{figure}[t]
\setlength{\abovecaptionskip}{-0.1cm}
\begin{center}
\includegraphics[width=1\linewidth]{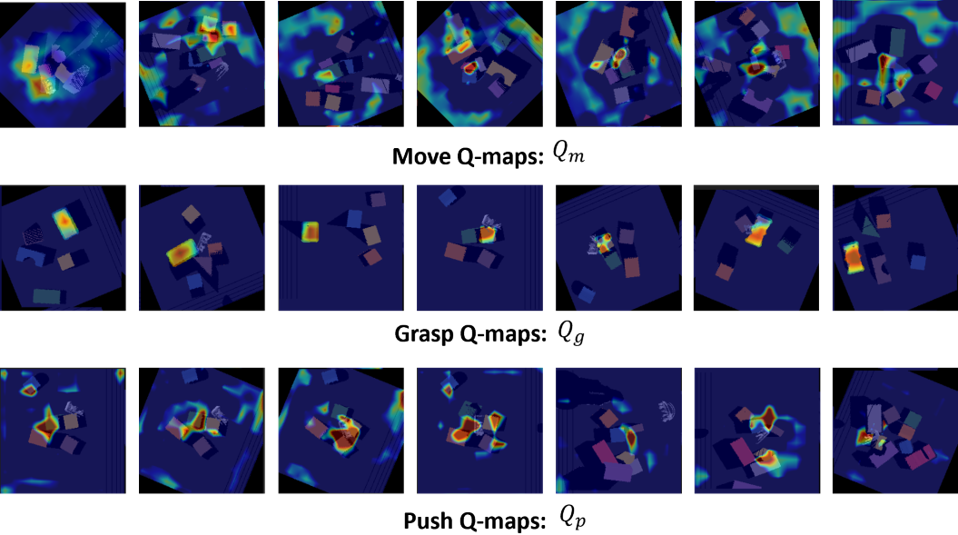}
\end{center}
\caption{Visualization of the Q-maps corresponding to the three primitive actions produced by MPGNet.}
\label{q-map}
\end{figure}

\begin{table}[t]\normalsize
    \centering
    \caption{Comparison of different methods in simulation}\label{tab_res}
    \resizebox{8.5cm}{!}{
    \begin{tabular}{p{4.2cm}<{\centering}|p{1.4cm}<{\centering}|p{1.4cm}<{\centering}|p{1.4cm}<{\centering}|p{1.4cm}<{\centering}}
        \hline
        \multirow{2}{*}[-1.5ex]{Method} & 
        \multicolumn{2}{c|}{Task success rate} & 
        \multicolumn{2}{c}{Average motion number} \\

        \cline{2-5} 
        & 15 objects & 30 objects & 15 objects & 30 objects \\
        \hline 
        Grasping-Only & 26\% & 12\% & 6.46 & 8.33 \\
        Target-Oriented VPG\cite{zeng2018learning} & 32\% & 26\% & 5.43 & 6.54 \\
        Grasping the Invisible\cite{yang2020deep} & 72\% & 66\% & 4.64 & 5.33 \\
        Efficient Push-Grasping\cite{xu2021efficient} & 84\% & 74\% & 3.88 & 4.97 \\
        \hline 
        MPGNet & 92\% & 84\% & 2.85 & 3.74 \\
        \hline 
    \end{tabular} }
\end{table}

\begin{figure}[t]
\setlength{\abovecaptionskip}{-0.1cm}
\begin{center}
\includegraphics[width=1\linewidth]{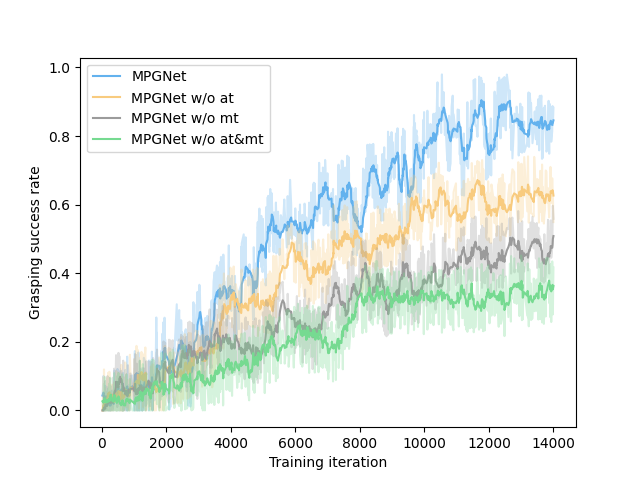}
\end{center}
\caption{Learning curves of MPGNet variants.}
\label{ablation study}
\end{figure}

\subsubsection{Number of Motions}
The number of motions is defined as the mean quantity of actions needed to effectively grasp the target, serving as an indicator of action efficiency.
In addition, it also shows the clutter of the scene as more complex scenes often require more actions to grasp the target object in general.
\subsubsection{Grasp Success Rate}
We record the average target grasping success rate for every $k$ training iterations to reflect the grasping capability of the model in the training process. It can be defined as the ratio of the number of successful target grasps to the number of grasp attempts.

\subsection{Simulation Experiments}
\begin{figure*}[t]
\setlength{\abovecaptionskip}{-0.1cm}
\begin{center}
\includegraphics[width=1\linewidth]{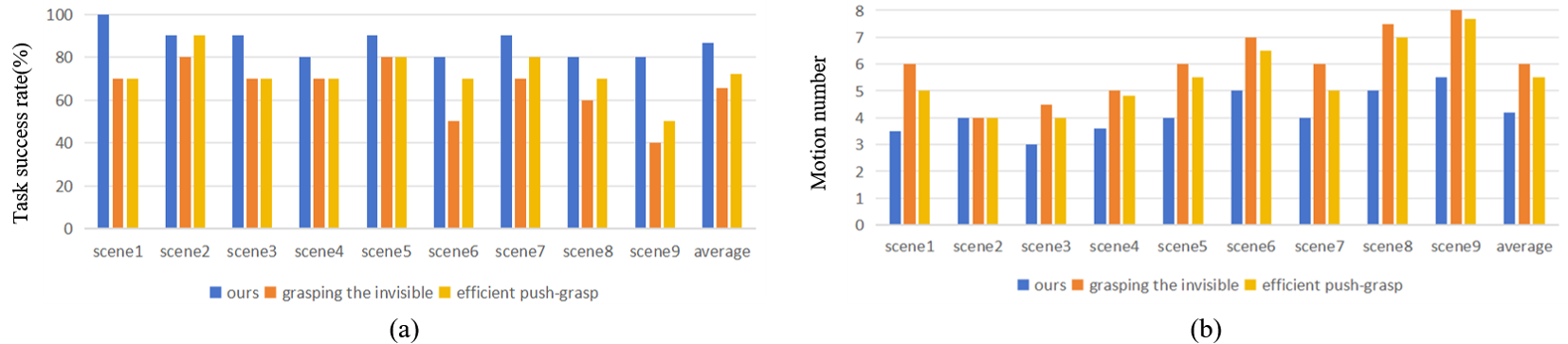}
\end{center}
\caption{Results of the real-world tests in terms of (a) the task success rate and (b) the average motion numbers for each test.}
\label{real-world test result}
\end{figure*}
We compare the performance of MPGNet with other baselines in both training and testing processes. 
In detail, Grasping the Invisible, Target-Oriented VPG method, Efficient Push-Grasping, and MPGNet have the same training scenario where we place $10$ objects in the workspace and the first or the second dropped object is the target object in the occluded scene.
For Grasping the Invisible and Target-Oriented VPG methods, we record the training process from scratch.
For Efficient Push-Grasping method, we start recording from the joint training.

We determine the mean success rate of target grasping after every 200 training iterations. During the training process, we limit the number of consecutive grasps to 2 for the methods except Grasping-Only. If this number exceeds 2 without grasping the target object successfully, the system will use a heuristic method to generate a grasp instead of the grasp-net to accelerate training. Also, if the number of total motions exceeds 10 without reaching the target, the scene will be reset. Compared with baselines (shown in Fig.~\ref{training performance}), the target grasping success rate of Grasping-Only still has no trend of convergence and stays lower than other methods. We can also see that the success rate of Target-Oriented VPG converges roughly to 0.35 similar to Grasping the Invisible, while the training process of the former one is not stable, indicating that the coordination policy in the latter method is slightly effective in the occlusion scenario. The curve of success rate shows that Efficient Push-Grasping (trained by the scheme proposed in \cite{xu2021efficient}) works better than the other three methods and the success rate convergence is approximately 0.6, which is still lower than the proposed method. For MPGNet, the training process shown in Fig.\ref{training performance} is recorded for stage V. It can be seen that MPGNet achieves the highest grasping success rate compared to other methods, which converges approximately to 0.9. 
It indicates that MPGNet works effectively and efficiently for target-oriented grasping in occlusion. The visualization of MPGNet's output is shown in Fig. \ref{q-map}.

We also conduct several test experiments in simulation to test whether our method can generalize to other occluded scenes.
We test two scenarios with 15 and 30 objects respectively. The test scenes are similar to the joint training scenarios while containing more objects. We randomly select one of the first five objects as the target to ensure that the target is highly occluded, aiming to show the robustness of MPGNet. We run 50 tests for each scenario. The results are reported in Table \ref{tab_res}, showing that our system achieves the best performance in terms of both task success rate and motion number.

\subsection{Ablation Study}
We create three variants of MPGNet, namely 
MPGNet w/o mt, MPGNet w/o at, and MPGNet w/o at\&mt. 
MPGNet w/o mt denotes that MPGNet is trained without the multi-stage strategy but with the alternative training, indicating that we begin training MPGNet from the joint training stage. MPGNet w/o at denotes that MPGNet is trained with the multi-stage strategy but without alternative training, indicating that the action decision in the joint training stage is based on the comparison of the Q values of the three action Q maps instead of the proposed alternative training scheme.
MPGNet w/o at\&mt denotes that the three action policies are trained jointly without any pre-training and action decisions are made by comparing the Q values.
Fig.\ref{ablation study} shows that MPGNet outperforms the other three ablated versions in terms of grasping success rate, which demonstrates the effectiveness of the multi-stage and the alternative training strategies.


\begin{figure}[t]
\setlength{\abovecaptionskip}{-0.1cm}
\begin{center}
\includegraphics[width=1\linewidth]{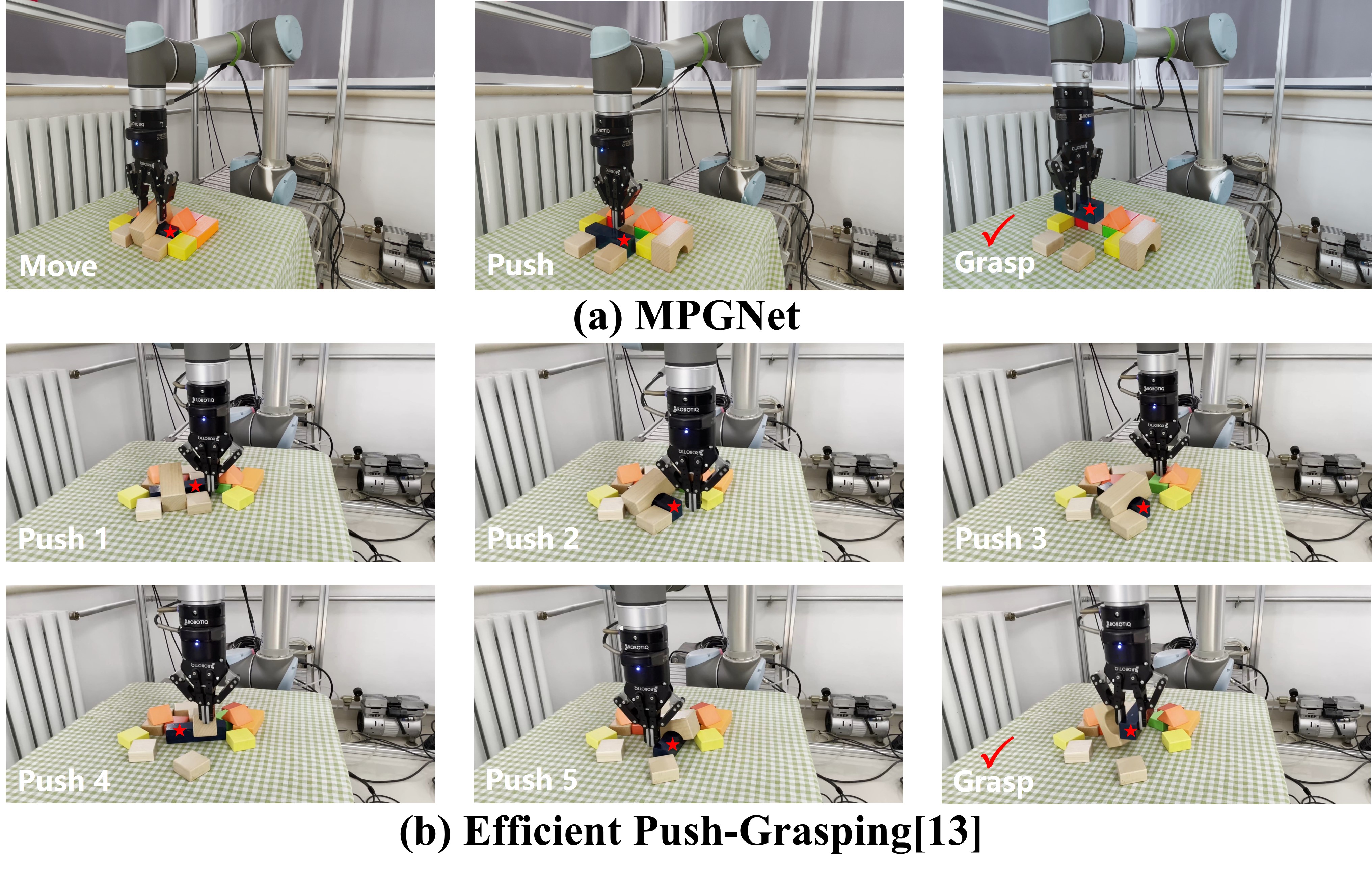}
\end{center}
\caption{Comparison of motion sequence. We compare MPGNet with Efficient Push-Grasping \cite{xu2021efficient} through real-world experiments. We can see that our method accomplishes the grasping task with fewer actions.}
\label{comparision of motion sequence}
\end{figure}

\subsection{Real-world Experiments}
In this section, we test our system in real-world scenarios. We conducted experiments in 9 challenging scenes with seen and unseen objects, in which the target objects are densely occluded. Each scenario contains objects with various textures and shapes. 
We evaluate our approach by contrasting it with the Efficient Push-Grasping and Grasping the Invisible, both of which have demonstrated commendable results in simulated evaluations.
For each test scene, we run 10 rounds. For each round of testing, we ensure consistency in the way the target objects were occluded and adjusted the placement of the remaining objects appropriately. The models we use in real-world experiments are transferred from simulation to reality without any fine-tuning. Results of task success rate and motion number are shown in Fig.\ref{real-world test result}. It can be seen that MPGNet outperforms the baselines across all the evaluation metrics. Fig.\ref{comparision of motion sequence} shows that our method can grasp the target object with the least motions.
As for the generation of the target mask heightmaps, we fine-tune the UOAIS \cite{back2022unseen} to our scene to generate target mask heightmaps. 
The output of UOAIS is a set of segmentation masks of the objects in the scene with occlusion labels. 
In practice, we choose the occluded object with the smallest value in height as the target to ensure that occlusion occurs. 
Furthermore, in some extra real-world scenarios, we introduce ChatGPT4 \cite{Gpt-4} as the multimodal large model with the proposed grasp algorithm to realize the occluded object grasping following the language instructions, making our real-world tests more realistic and practical. In detail, we query GPT4 to categorize the detected masks, and the candidate categories are given in advance. Then we prompt the GPT4 to analyze the input language instructions and select the target mask as input for the proposed algorithm.
Video demonstration of the real-world tests can be found at: \url{https://youtu.be/S_QKZqkh0w8}.



\section{Conclusions}

In this work, we propose MPGNet, a three-branch network to learn moving, pushing, and grasping synergy for target-oriented grasping in occluded scenes. In practice, we train our network via a multi-stage strategy and combine domain knowledge with Q values to better coordinate the three actions. We compare our method with several baselines, demonstrating that MPGNet converges rapidly with a high grasping success rate. We also evaluate MPGNet via both simulation and real-world experiments, showing that it works well challenging scenes including those that can not be efficiently handled by common push-grasping synergy. Moreover, we demonstrate that MPGNet can be assisted with ChatGPT4 to achieve instruction-based target object grasping in occluded scenes. 


\addtolength{\textheight}{-12cm}   







\bibliographystyle{ieeetr}
\bibliography{my_bib}
\end{document}